\documentclass[10pt, conference, compsocconf]{IEEEtran}

\usepackage{cite}
\usepackage[pdftex]{graphicx}
\usepackage{multirow}

\usepackage[cmex10]{amsmath}
\usepackage{amssymb}
\usepackage[font=footnotesize,caption=false]{subfig}
\usepackage{url}
\usepackage{etoolbox}
\apptocmd{\thebibliography}{\scriptsize}{}{}

\newcommand{\squeezeup}{\vspace{-2.5mm}}

\usepackage[super]{nth}

\begin{document}

\title{Document Image Binarization with Fully Convolutional Neural Networks}

\author{\IEEEauthorblockN{Chris Tensmeyer and Tony Martinez}
\IEEEauthorblockA{Dept. of Computer Science\\
Brigham Young University\\
Provo, USA\\
tensmeyer@byu.edu martinez@cs.byu.edu}
}


\maketitle

\begin{abstract}
Binarization of degraded historical manuscript images is an important pre-processing step for many document processing tasks.
We formulate binarization as a pixel classification learning task and apply a novel Fully Convolutional Network (FCN) architecture that operates at multiple image scales, including full resolution.
The FCN is trained to optimize a continuous version of the Pseudo F-measure metric and an ensemble of FCNs outperform the competition winners on 4 of 7 DIBCO competitions.
This same binarization technique can also be applied to different domains such as Palm Leaf Manuscripts with good performance.
We analyze the performance of the proposed model w.r.t. the architectural hyperparameters, size and diversity of training data, and the input features chosen.

\end{abstract}


\section{Introduction}

To minimize the impact of physical document degradation on document image analysis tasks (e.g. page segmentation, OCR), it is often desirable to first binarize the digital images.
Such degradations include non-uniform background, stains, faded ink, ink bleeding through the page, and un-even illumination.
Binarization separates the raw content of the document from these noise factors by labeling each pixel as either foreground ink or background.
This is a well-studied problem, even for highly degraded documents, as evidenced by the popularity of the Document Image Binarization Content (DIBCO) competitions~\cite{dibco09,hdibco10,dibco11,hdibco12,dibco13,hdibco14,hdibco16}.

We present a learning model for document image binarization.
Fully Convolutional Networks (FCN)~\cite{long15} alternate convolution and non-linear operations to efficiently classify all pixels of an input image in a single forward pass.
While FCNs have been previously proposed for semantic segmentation tasks, their output predictions are poorly localized due to high downsampling.
We propose a multi-scale architectural variation that results in precise localization and preserves the generalization benefits of downsampling.

FCNs can be applied to diverse domains of documents images without tuning hyperparameters.
For example, we show that the same FCN architecture can be trained to achieve state-of-the-art performance on both historical paper documents (i.e. DIBCO data) and on palm leaf manuscripts.
Current state-of-the-art methods for paper documents (e.g.\cite{howe13}) tend to perform poorly on palm leaf manuscripts due to domain differences between the tasks~\cite{burie16}.
FCNs also automatically adapt to any particular definition of binarization ground truth (e.g. see ~\cite{smith12}) because they do not incorporate explicit prior information.

Many common binarization approachs compute local or global thresholds based on image statistics~\cite{otsu75,sauvola00}.
One disadvantage of this approach is the threshold is invariant to a permutation of the pixels, i.e., statistics ignore shape.
On the other hand, other approaches (e.g. edge detection, Markov Random Field (MRF), connected components) include strong biases about the shape of foreground components.
In contrast, FCNs learn from training data to exploit the spatial arrangements of pixels without relying on a hand-crafted bias on local shapes.

Previous learning approaches are trained to optimize per-pixel accuracy, which is problematic due to the majority of pixels being background.
These methods typically resort to heuristic sampling of pixels to achieve a balanced training set.
Instead, we propose directly optimizing a continuous adaptation of the Pseudo F-measure (P-FM)~\cite{ntirogiannis13} which has been a DIBCO evaluation metric since 2013~\cite{dibco13}.
Because P-FM does not penalize foreground border pixels that are predicted as background, we combine P-FM with regular F-measure (FM) so the FCN correctly classifies border pixels.

The contributions of this work are as follows.
First, we propose the use of FCNs for document image binarization and determine good architectures for this task.
We show that directly optimizing the proposed continuous Pseudo F-measure exceeds the previous state-of-the-art on DIBCO competition data.
We compute a learning curve and show that diversity of data is more important than quantity of data.
Finally, we demonstrate that FCN performance can be improved though additional input features, including the output of other binarization algorithms.

\section{Related Work}

\begin{figure*}[t]
\centering
\includegraphics[height=2in]{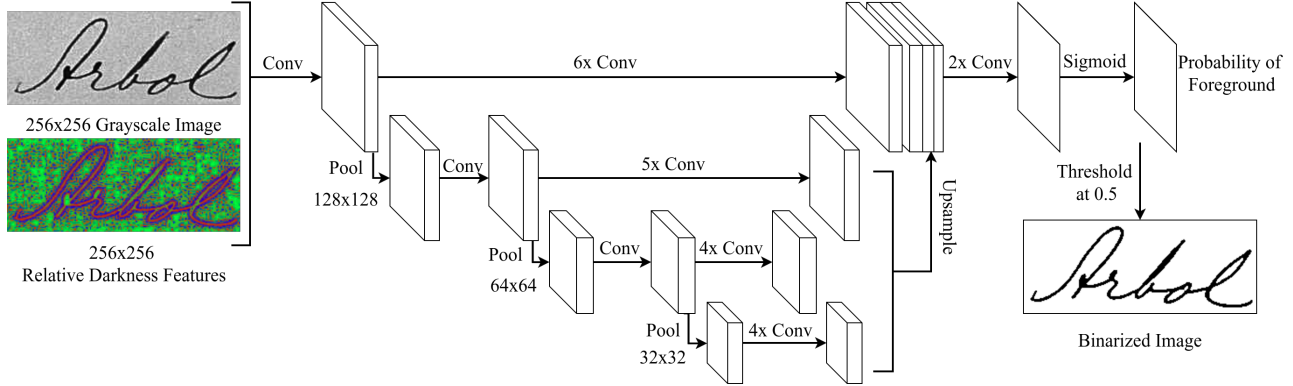}
\caption{Multi-Scale Fully Convolutional Neural Network architecture.  All Conv operations are followed by ReLU, and ``Nx Conv'' labels indicate N successive conv layers.  All pool layers downsample by a factor of 2.  Upsampling is done by bilinear interpolation to the input spatial dimensions.}
\label{fig:fcn_branch}
\squeezeup
\squeezeup
\end{figure*}

Howe's method formulates image binarization as energy minimization over a MRF~\cite{howe13}.
The unary energy terms are computed from the image Laplacian and the pairwise connections are determined by Canny Edge detection~\cite{canny86}.
An exact minimization of the energy function can be obtained by solving the equivalent Max Flow problem~\cite{boykov04}.
Variants of this method have placed \nth{1} in HDIBCO 2012, 2014, 2016 and \nth{2} in DIBCO 2013~\cite{hdibco12,hdibco14,hdibco16,dibco13}.
The DIBCO 2013 winner combines local image contrast and local image gradients to determine edges between text and background~\cite{su13}.

Several classification-based binarization approaches have been proposed.
Hamza et al. train a Multi-Layer Perceptron (MLP) classifier on pixel labels derived from clustering~\cite{hamza05}.
Kefali et al. train an MLP using ground truth data to classify each pixel based on the intensity values of its 3x3 neighborhood and the global image mean and standard deviation~\cite{kefali14}, though they do not out perform Sauvola's method~\cite{sauvola00}.
Afzal et al. trained a 2D Long Short Term Memory (LSTM) network which incorporates both local and global information~\cite{afzal15}.
Their approach greatly reduced OCR error on the binarized text compared to Sauvola.

Pastor-Pellicer et al. explored the use of Convolutional Neural Networks (CNN) to classify each pixel given its 19x19 neighborhood of intesity values~\cite{pastor15}.
They report an FM of 87.74 on DIBCO 2013 compared to 92.70 achieved by the competition winner.
Wu et al. trained an Extremely Randomized Trees classifier using a wide variety of statistical and heuristic features extracted from various neighborhoods around the pixel of interest.
Because the number of background pixels greatly exceeds the number of foreground pixels, they heuristically sampled a training set to balance both classes.
In contrast, we directly optimize the Pseudo F-measure instead of determining the precision/recall tradeoff through sampling.

Long et al. proposed FCNs for the more general semantic segmentation problem in natural images~\cite{long15}.
Both Zheng et al. and Chen et al. combined Conditional Random Fields (CRF) with FCNS to improve localization and consistency of predictions~\cite{zheng15,chen14}.
These FCNs heavily downsample inputs, which results in poor localization.
Thus prior FCNs are not good models for document image binarization.

\section{Methods}

In this section, we describe FCNs, our multi-scale extension, and the proposed loss function we use to train them.

\subsection{Fully Convolutional Networks}

Fully Convolutional Networks (FCN) are models composed of alternating convolution operations with element-wise non-linearities.
The FCNs we consider map an input image $x \in \mathbb{R}^{D \times H \times W} \rightarrow y \in \mathbb{R}^{H \times W}$, where $y_{ij} \in [0,1]$ is the probability that pixel $x_{ij}$ is foreground, and $D$ is the number of input channels (e.g. 3 for RGB).

Specifically, the  $\ell{}^{th}$ layer ($1 \leq \ell \leq L$) of the FCN performs a convolution operation with learnable kernels followed by a rectification:
\begin{equation} \label{eq:fcn}
x_{\ell} = \operatorname{ReLU}(W_{\ell} \ast x_{\ell - 1} + b_{\ell})
\end{equation}
where $x_{\ell} \in \mathbb{R}^{ D_{\ell} \times H \times W }$ is the output of layer $\ell$, $x_0$ is the input image, $\operatorname{ReLU}(z) = \max(0, z)$ is element-wise rectification, $W_{\ell} \in \mathbb{R}^{D_{\ell} \times K_{\ell} \times K_{\ell} \times D_{\ell - 1}}$ are convolution kernels, and $b_{\ell} \in \mathbb{R}^{D_{\ell}}$ is a bias term for each kernel.
To obtain probabilities for each pixel, a sigmoid function is applied element-wise to $x_L \in \mathbb{R}^{1 \times H \times W}$ without rectification.

Each $W_{\ell}$ can be interpreted as $D_{\ell}$ distinct convolution kernels (i.e. $W_{\ell,i} \in \mathbb{R}^{K_{\ell} \times K_{\ell} \times D_{\ell - 1}}$) that have two spatial dimensions, each of size $K_{\ell}$, and a depth dimension of size $D_{\ell - 1}$, which is the number of kernels used in the previous layer.
Thus convolution is carried out in two spatial dimensions, but each $W_{\ell,i}$ spans all channels of its layer's input.
Each of these kernels is applied independently to $x_{\ell - 1}$ to yield a single channel $x_{\ell,i} \in \mathbb{R}^{H \times W}$.
Each $x_{\ell,i}$ then becomes a single channel of the multi-channel $x_{\ell}$.

In our paper, we simplify the choice of hyper-parameters by using the same value for many parameters, regardless of $\ell$.
For example, we primarily use $D_{\ell}=64$ and $K_{\ell}=9$.

\subsection{Multi-Scale}

Many applications of FCNs to dense pixel prediction problems find that incorporating information at multiple scales leads to significantly improved performance.
This is an elegant way to fuse local and global features for classification.
For example, the original FCN~\cite{long15} downsamples images to $\frac{1}{32}$ the input size and finds improved performance by incorporating features computed at scales $\frac{1}{8}$ and $\frac{1}{16}$.
In contrast, binarizing text images requires precise localization which is difficult when images are downsampled.

We utilize a branching FCN to compute features over scales $\frac{1}{1}$, $\frac{1}{2}$, $\frac{1}{4}$, $\frac{1}{8}$.
This is shown in Figure~\ref{fig:fcn_branch}.
This network has depth $L=9$, width $D_{\ell}=64$, 4 scales, and kernel size $K_{\ell}=9$.
At certain layers, 2x2 average pooling is applied to produce an additional branch of the FCN at a smaller scale.
After several convolution layers at each scale, the output of each scale is upsampled to the original size using bilinear interpolation.
This is followed by two more convolution layers applied to the concatenated scale outputs.
This allows the FCN to fuse both local and increasingly global features to classify pixels.


\subsection{Pseudo F-measure Loss}

\label{sec:p_fm}

Pseudo F-measure (P-FM) was formulated with the intuition that binarization errors on textual images should be penalized based on how they might obscure individual characters~\cite{ntirogiannis13}.
For example, false positives (actual background predicted as foreground) far from foreground components have a smaller penalty than false positives between characters.
P-FM is the harmonic mean of Pseudo Recall and Pseudo Precision, which we denote $R_{ps}$ and $P_{ps}$ respectively.
These quantities are computed as
\begin{equation} \label{eq:p_fm}
F_{ps} = \frac{2 R_{ps} P_{ps}}{R_{ps} + P_{ps}}
\end{equation}
\begin{equation} \label{eq:p_recall}
R_{ps} = \frac{\Sigma_{xy} (B_{xy} G_{xy} R^W_{xy})}{\Sigma_{xy} (G_{xy} R^W_{xy})}
\end{equation}
\begin{equation} \label{eq:p_precision}
P_{ps} = \frac{\Sigma_{xy} (G_{xy} B_{xy} P^W_{xy})}{\Sigma_{xy} (B_{xy} P^W_{xy})}
\end{equation}
where $B_{xy}$ is predicted probability of foreground for the pixel $(x,y)$, $G_{xy} \in \{0,1\}$ is the ground truth class of pixel $(x,y)$, and $R^W$ and $P^W$ are per-pixel weights for recall and precision errors respectively.

While $R^W$ and $P^W$ can be arbitrarily specified, we compute them using~\cite{ntirogiannis13} to yield the P-FM used in the DIBCO evaluation protocol.
Using uniform weights yields the more common F-measure, which was explored in \cite{pastor13} as the training objective of neural networks.

To use Eq.~\ref{eq:p_fm} as the objective function in Stochastic Gradient Descent (SGD) training of the FCN, we must derive an expression for $\frac{d F}{d B}$.  For simplicity, we drop the subscripts on $F_{ps}, R_{ps}, P_{ps}$.
Taking the derivative of Eq.~\ref{eq:p_fm}, we have
\begin{equation}
\frac{dF}{dB} = 
\frac{\partial F}{\partial P} \frac{dP}{dB} + \frac{\partial F}{\partial R} \frac{dR}{dB} =
2\frac{ \frac{dR}{dB} P^2 + \frac{dP}{dB} R^2}{(P + R)^2}
\end{equation}
Noting that $\frac{dR}{dB}$ is a matrix of partial derivatives, deriving $\frac{\partial R}{\partial B_{ij}}$ is simple as $B$ only occurs in the numerator in Eq.~\ref{eq:p_recall}.
\begin{equation}
\frac{\partial R}{\partial B_{ij}} = 
\frac{\partial}{\partial B_{ij}} \bigg[ \frac{\Sigma_{xy} (B_{xy} G_{xy} R^W_{xy})}{\Sigma_{xy} (G_{xy} R^W_{xy})} \bigg] =
\frac{G_{ij}R^W_{ij}}{\Sigma_{xy} (G_{xy} R^W_{xy})}
\end{equation}
Similarly, we can derive $\frac{\partial P}{\partial B_{ij}}$ by applying the quotient rule for derivatives to Eq.~\ref{eq:p_precision} and simplifying.
\begin{equation}
\frac{\partial P}{\partial B_{ij}} = 
P^W_{ij} \frac{\Sigma_{xy} (B_{xy} P^W_{xy}) G_{ij} - \Sigma_{xy} (G_{xy} B_{xy} P^W_{xy})}{[\Sigma_{xy} (B_{xy} P^W_{xy})]^2}
\end{equation} 

\subsection{Datasets and Metrics}

We primarily use two datasets for evaluation: DIBCOs~\cite{dibco09,hdibco10,dibco11,hdibco12,dibco13,hdibco14,hdibco16} and Palm Leaf Manuscripts (PLM)~\cite{burie16}.
For DIBCO experiments, we used all 86 competition images from the years 2009-2016 for either training, validation, or test data.
When testing on a particular DIBCO year, that year's competition data composes the test set, 10 random other images compose the validation set, and the remaining images compose the training set.
For PLM, we randomly split the 50 training images into 40 training and 10 validation and use the designated 50 images for testing~\cite{burie16}.
While there are two sets of ground truth annotations for PLM, for simplicity we use only the first set.

To avoid overfitting of testing data, development of our method was performed using the HDIBCO 2016 validation set.
The test data was only used in the final evaluation and not used to select models, architectures, or features.

The metrics we report are from the DIBCO 2013 evaluation criteria: P-FM, FM, Peak Signal to Noise Ratio (PSNR), and Distance Reciprocal Distortion (DRD).
For P-FM, FM, and PSNR, higher numbers indicate better performance.
For DRD, lower is better.

\subsection{Implementation Details}

\begin{figure*}

\subfloat[Image]{\includegraphics[width=0.237\textwidth]{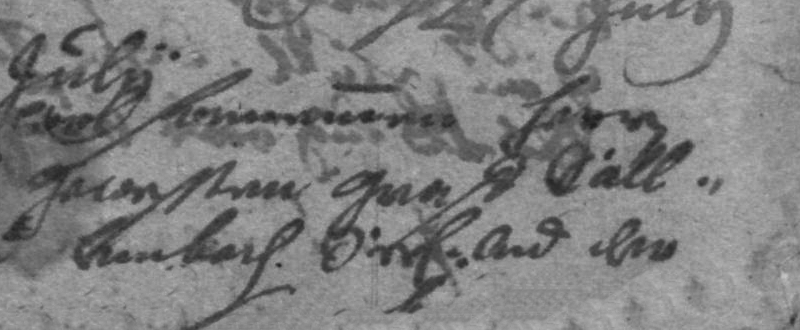}} \hspace{5pt}
\subfloat[Ground Truth]{\includegraphics[width=0.237\textwidth]{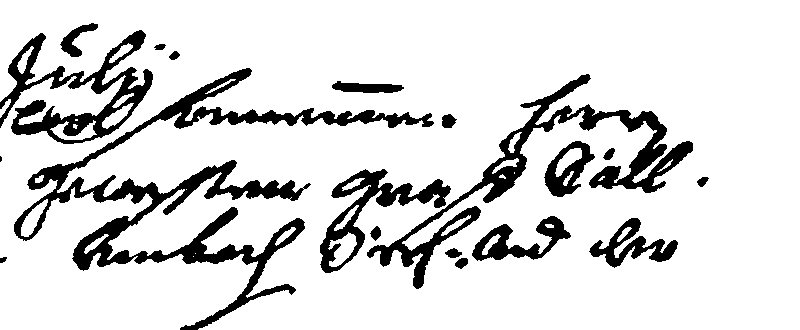}} \hspace{5pt}
\subfloat[Howe Binarization]{\includegraphics[width=0.237\textwidth]{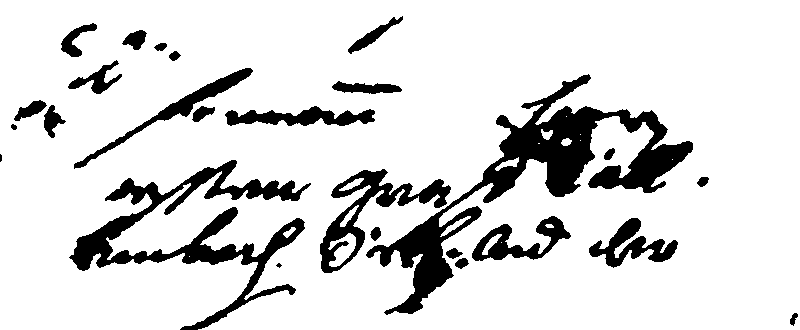}} \hspace{5pt}
\subfloat[Proposed FCN Binarization]{\includegraphics[width=0.237\textwidth]{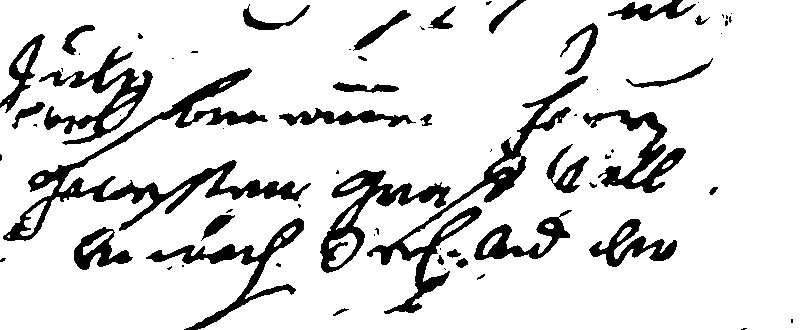}}

\subfloat[Image]{\includegraphics[width=0.237\textwidth]{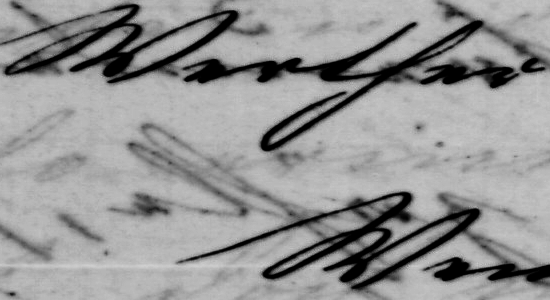}} \hspace{5pt}
\subfloat[Ground Truth]{\includegraphics[width=0.237\textwidth]{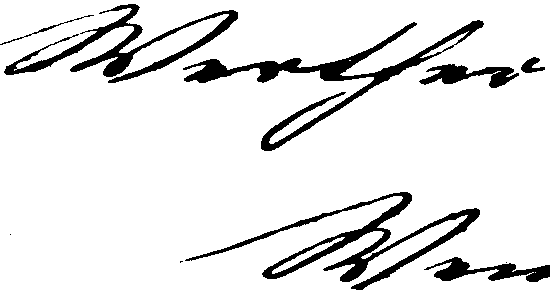}} \hspace{5pt}
\subfloat[Howe Binarization]{\includegraphics[width=0.237\textwidth]{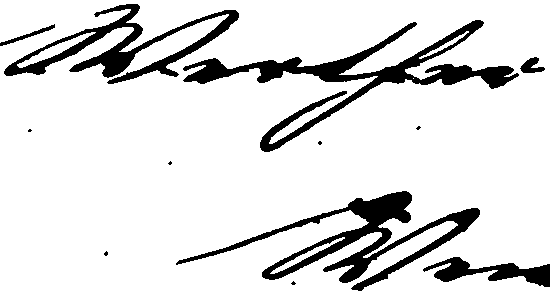}} \hspace{5pt}
\subfloat[Proposed FCN Binarization]{\includegraphics[width=0.237\textwidth]{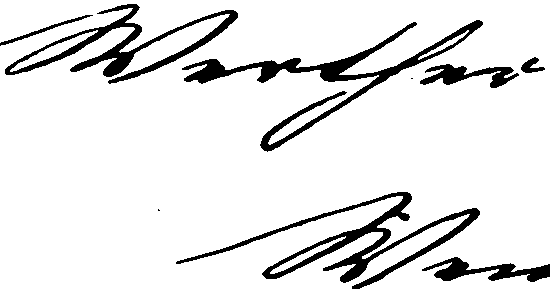}}

\caption{Qualitative comparison of proposed ensemble of FCNs with state-of-the-art Howe Binarizataion~\cite{howe13}.  Images contain  significant bleed through noise and come from the H-DIBCO 2016 test data. }
\label{fig:compare}
\end{figure*}

\begin{table*}[t]
\centering
\begin{tabular}{|c|l|c|c|c|c|}
\hline
    &             & \multicolumn{4}{c|}{Metrics} \\
\cline{3-6}
Dataset & Loss             & P-FM                   & FM                     & DRD                  & PSNR  \\
\hline
\multirow{4}{*}{HDIBCO 2016}
&P-FM             & \textbf{94.09 (94.67)} & 86.66 (87.06)          & 4.62 (4.38)          & 17.73 (17.86) \\
&FM               & 92.90 (93.23)          & 89.93 (90.30)          & 3.69 (3.51)          & \textbf{18.73 (18.90)} \\
&P-FM + FM        & 93.22 (93.76)          & 89.01 (89.52)          & 4.01 (3.76)          & 18.48 (18.67) \\ 
&Cross-Entropy    & 92.59 (92.94)          & \textbf{90.20 (90.56)} & \textbf{3.62 (3.45)} & 18.68 (18.84)  \\
\hline
\multirow{4}{*}{PLM}
&P-FM             & 68.23 (68.55)          & 66.93 (67.20)          & 9.24 (9.10)           & 14.79 (14.83) \\
&FM               & 67.40 (67.74)          & \textbf{68.38 (68.69)} & 9.86 (9.68)           & 14.59 (14.64) \\
&P-FM + FM        & \textbf{68.54 (68.96)} & 68.27 (68.63)          & \textbf{9.12 (8.94)}  & \textbf{14.81 (14.87)} \\
&Cross-Entropy    & 66.41 (66.77)          & 65.38 (65.68)          & 9.95 (9.78)           & 14.58 (14.63) \\
\hline



\end{tabular}
\caption{Average performance of 5 FCNs on H-DIBCO 2016 and PLM datasets for various loss functions.  Numbers in parenthesis indicate ensemble performance.}
\label{tab:loss}
\squeezeup
\squeezeup
\squeezeup
\end{table*}

We used the deep learning library Caffe for all experiments. 
Our training and validation sets are composed of 256x256 crops extracted at a stride of 64 pixels from the input images.
Image crops composed of only background pixels are discarded as Pseudo Recall is undefined for all background predictions.
Our FCNs take as input the gray scale image plus locally computed Relative Darkness features (see Section~\ref{sec:features}).

To binarize a whole image, the FCN binarizes individual overlapping 256x256 crops and extracts the center 128x128 patch of each output crop (except for border regions).
This ensures each pixel has sufficient context for the FCN to classify it.
These patches are stitched together to form the whole binarized image.

We used SGD to train FCNs with an initial learning rate (LR) of 0.001, mini-batch size of 10 patches, and an L2 weight decay of 0.0005.
Gradient clipping to an L2 norm of 10 was employed to help stablize learning~\cite{pascanu13}.
The LR was multiplied by 0.1 if performance on the validation set failed to improve for 1.5 epochs.
Training ended when the LR decayed to $10^{-6}$ and the set of parameters that best performed on the validation set is the output of the training procedure.
During training, we stochastically add a small constant to each channel of the input image as color jitter, which empirically improves performance by a small amount.

\section{Experiments}

In this section, we give results and discussion for our experiments to validate the proposed model.

\subsection{Loss Functions}

In this set of experiments, we compare 4 loss functions on HDIBCO 2016 and PLM: P-FM, FM, P-FM + FM, and Cross Entropy (CE).
Because the P-FM loss function does not penalize the outer border of foreground components being classified as background (due to recall weights of 0), we experimented with adding together P-FM and FM losses during training.
CE loss is a standard classification based loss that optimizes per-pixel accuracy.

For each loss function, we trained 5 FCNs from different random initializations.
Table~\ref{tab:loss} shows both the average performance of individual FCNs as well as the combined ensemble performance in parenthesis.
The outputs of individual FCNs are combined using per-pixel majority vote.
This improves performance by a significant amount at the cost of increased computation.

Unsurprisingly, optimizing for P-FM (or P-FM + FM) yields the best performance for P-FM, with CE performing worst for P-FM.
Training with only P-FM lowers performance in other metrics due to predicting border pixels as background.
While developing our method, FM loss generally out-performed CE loss on all metrics on validation data, though surprisingly CE performed better on the HDIBCO 2016 test data.
Nevertheless, based on our validation set performance, we chose to use P-FM + FM loss for the remaining experiments.
For reference, the best published FM on PLM is 68.76~\cite{burie16}, which was achieved by a single scale FCN that was pretrained on DIBCO and proprietary data.

Figure~\ref{fig:compare} shows a qualitative comparison between Howe's method~\cite{howe13} (using the author's code) and the ensemble of 5 FCNs trained with P-FM + FM loss.
Because Howe's method uses pixel connectivity, it sometimes misclassifies small background regions as foreground when they are bordered by both foreground ink and bleed-through noise.
On the other hand, FCNs compute local features on multiple scales, so it learns to recognize whether text is bleed-through noise or actual foreground ink.

\subsection{DIBCO performance}
\begin{table*}[t]
\centering
\begin{tabular}{|l|c|c|c|c||c|c|c|c|}
\hline
& \multicolumn{4}{c||}{Proposed Method} & \multicolumn{4}{c|}{Best Competition System} \\
\cline{2-9}
Competition      & P-FM  & FM    & DRD  & PSNR  & P-FM  & FM    & DRD  & PSNR  \\
\hline
DIBCO 2009       & 92.59 & 89.76 & 4.89 & 18.43 &  -    & \textbf{91.24} & -    & \textbf{18.66} \\
H-DIBCO 2010     & 97.65 & \textbf{94.89} & 1.26 & \textbf{21.84} &  -    & 91.50 & -    & 19.78 \\
DIBCO 2011       & 97.70 & \textbf{93.60} & \textbf{1.85} & \textbf{20.11}  &  -    & 88.74 & 5.36 & 17.97 \\
H-DIBCO 2012     & 96.67 & 92.53 & \textbf{2.48} & 20.60 &  -    & \textbf{92.85} & 2.66 & \textbf{21.80} \\
DIBCO 2013       & \textbf{96.81} & \textbf{93.17} & \textbf{2.21} & 20.71 & 94.19 & 92.70 & 3.10 & \textbf{21.29} \\
H-DIBCO 2014     & 94.78 & 91.96 & 2.72 & 20.76 & \textbf{97.65} & \textbf{96.88} & \textbf{0.90} & \textbf{22.66} \\
H-DIBCO 2016     & \textbf{93.76} & \textbf{89.52} & \textbf{3.76} & \textbf{18.67} & 91.84 & 88.72 & 3.86 & 18.45 \\
\hline

\end{tabular}
\caption{Comparison of ensemble of 5 FCNs with DIBCO competition winners.  The ``-" symbols indicate unreported metrics.}
\label{tab:hdbico_all}
\squeezeup
\squeezeup
\squeezeup
\end{table*}

Here we compare our ensemble of 5 FCNs with the best performing competition submissions for each of the 7 DIBCO competitions.
The winning systems did not necessarily have the best performance wrt all metrics.
Therefore, we compare with the best score per metric for any system entered in the competition.
For example, in DIBCO 2013, the rank 1 system had the highest P-FM (94.19), but the rank 2 system had the highest FM (92.70).

\begin{figure}
\centering
\subfloat[Image]{\includegraphics[width=0.23\textwidth]{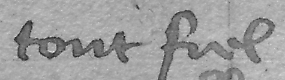}} \hspace{5pt}
\subfloat[Ground Truth]{\includegraphics[width=0.23\textwidth]{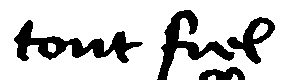}}

\subfloat[Proposed Binarization]{\includegraphics[width=0.23\textwidth]{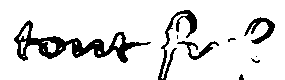}}
\caption{Failure case on H-DIBCO 2014.  None of the training images are similar, so FCN generalizes poorly to this kind of ink.}
\label{fig:failure}
\squeezeup
\end{figure}

The results are presented in Table~\ref{tab:hdbico_all}.
For 4 competitions, including the most recent HDIBCO 2016, the ensemble of FCNs out-performs the best submitted entries.
Our FM is low on DIBCO 2009 due to false positives located far away from the foreground ink.
Under the P-FM, these mistakes count for less because they don't obscure the readability of the text.
For HDIBCO 2014, our method performs very poorly on image 6 (see Figure~\ref{fig:failure}), where it achieves a P-FM of 56.4.
On the other 9 H-DIBCO 2014 images, it achieves almost perfect performance with an average P-FM of 99.04.

\subsection{Architecture Search}

\begin{figure}
\centering
\includegraphics[width=0.48\textwidth]{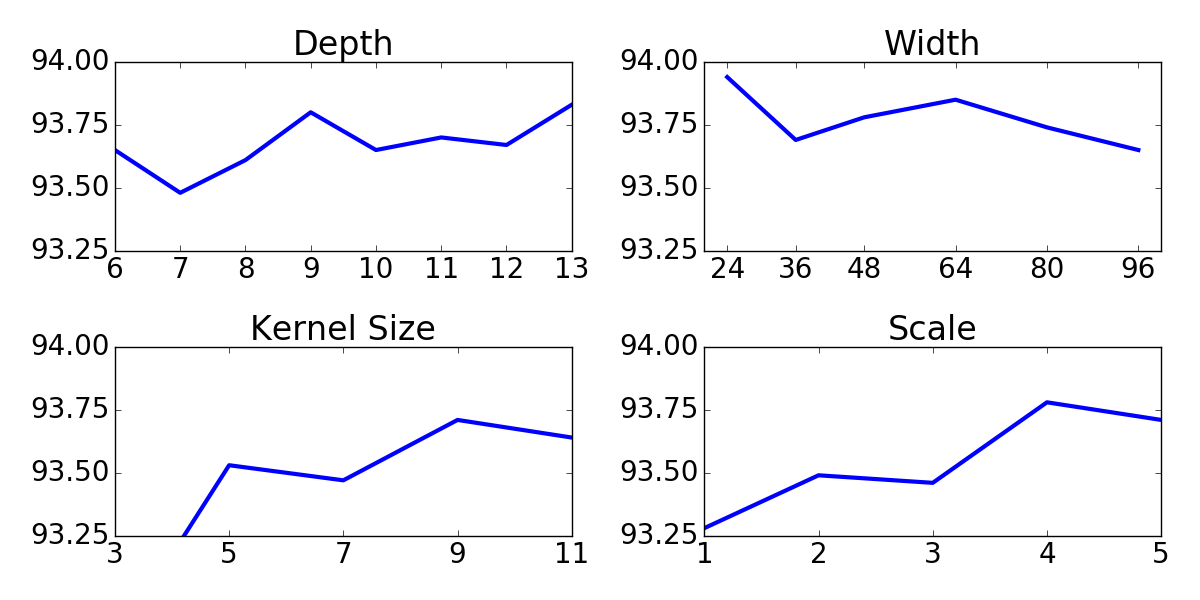}

\caption{P-FM on HDIBCO 2016 as a function of architecture hyperparameters evaluated with an ensemble of 5 FCNs.  In general, performance is not sensitive to any particular combination of hyperparameters.}
\label{fig:arch}
\squeezeup
\end{figure}

FCNs have a number of hyperparameters that together constitute the architecture of the network.
Here we independently vary four architecture components and measure P-FM on the HDIBCO 2016 dataset.
Specifically, we vary network depth, width, number of scales, and kernel size.
The base architecture (used in previous experiments) is depth $L=9$, width $D_{\ell}=64$, 4 scales, and kernel size $K_{\ell}=9$.

The results for ensembles of 5 FCN for each architecture are presented in Figure~\ref{fig:arch}.
Performance seems relatively insensitive to architectural hyperparameters, with the exception that kernel sizes of 3 perform worse than larger kernels.
Improvement due to ensemble prediction is typically larger than performance differences among architectures.

\subsection{How Much Data is Enough?}

\begin{figure}
\centering
\includegraphics[width=0.4\textwidth]{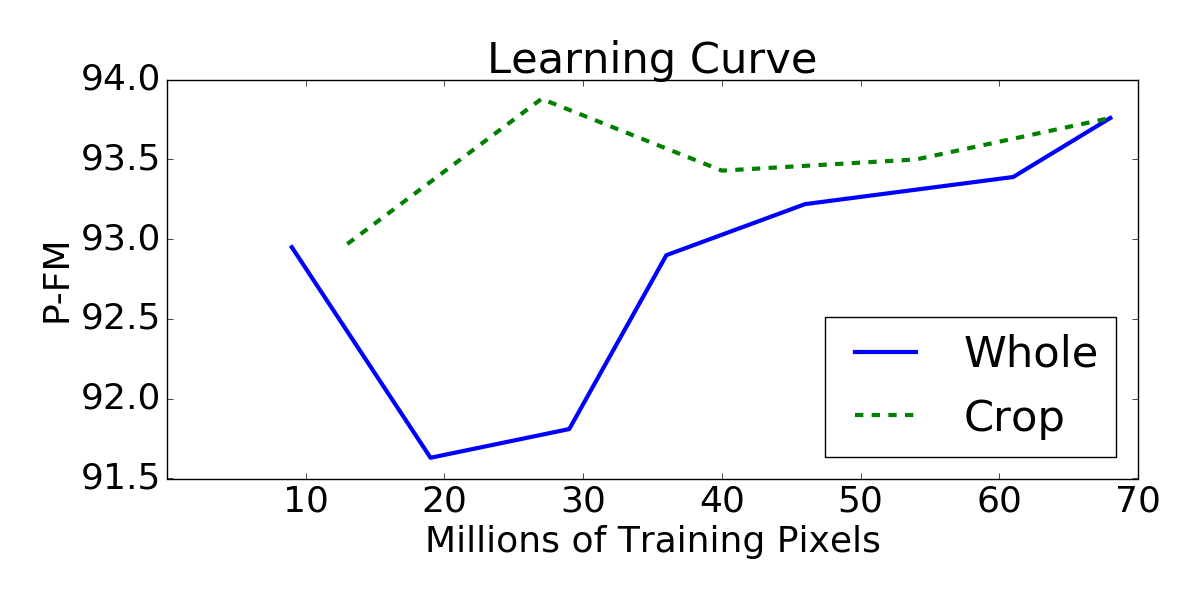}

\caption{Learning curve for two methods of shrinking the training data.  The solid blue line indicates that whole images were removed from the training set.  The dashed green line indicates that all training images were kept, but were cropped to be smaller.}
\label{fig:learning_curve}
\squeezeup

\end{figure}

More data typically leads to better performing learning models, though eventually adding more data yields diminishing returns.
Here we analyze the amount of data need by constructing a learning curve with amount of data on the x-axis and P-FM on the y-axis for the HDIBCO 2016 dataset.
Figure~\ref{fig:learning_curve} shows the curve for two different ways of varying the amount of training data (keeping the validation data the same).
The first way simply reduces the number of training images along the x-axis.
This decreases the diversity of images that compose the training set.
The second way retains all 76 training/validation images, but crops each image to be 20\%, 40\%, 60\%, or 80\% of its original width.
This also decreases the number of pixels for training, but retains diversity of inputs presented to the FCN.

Even at similar number of pixels, cropping training images instead of removing them performs better and indicates that greater diversity in the training data is a key factor for improving performance.
This is logical because if an image is homogeneous in handwriting style and noise content, then many of the pixels are locally similar.
Thus, if the goal of manual or semi-automatic creation of ground truth binarizations is to create more training data, the annotator should focus on small, diverse images instead of fully annotating large images.

Training on 10M pixels resulted in better test set performance than on 20-40M pixels for whole images.
We find this as evidence of overfitting the validation set, which is used for model selection.
The validation P-FM for 10M and 20M pixels are 92.84 and 95.26 respectively.
Additionally, the added training images are more similar to the 10 validation images than the 10 test images.
This shows that there is significant dataset bias in evaluation with so few number of images, even if the number of pixels is very large.


\subsection{Input Features}

\label{sec:features}

\begin{table}
\centering
\begin{tabular}{|c|c|c||c|c|c|}
\hline
Extra Feature       & Size       & P-FM & Extra Feature       & Size        & P-FM  \\
\hline
None                & -          & 96.63 & \nth{10} Percentile & 3          & 96.69 \\
Min                 & 9          & 96.73 & \nth{25} Percentile & 3          & 96.63 \\
Max                 & 3          & 96.74 & Canny Edges         & -          & 96.93 \\
Median              & 39         & 96.57 & Percentile Filter   & -          & 96.16 \\
Mean                & 39         & 96.38 & Bilateral Filter    & 100        & 96.01 \\
Otsu                & -          & 95.55 & Standard Deviation             & 3          & 95.65 \\
Howe                & -          & 97.14 & Relative Darkness   & 5          & \textbf{97.15} \\

\hline
\end{tabular}

\caption{P-FM on HDIBCO 2016 by including additional input features. The ``-" symbols indicate globally computed features.}
\label{tab:features}
\squeezeup
\squeezeup
\end{table}

While FCNs can learn good discriminative features from raw pixel intensities, there may exist useful features that the FCN is not able to learn because they cannot be efficiently approximated by alternating convolution and rectification operations.
For example, Sauvola's method uses local standard deviation which is not easy for an FCN to learn.

We experimented with input features that are densely computed and treated as additional input image channels.
For this experiment, we used a batch size of 5 and minimized P-FM loss.
We report P-FM on HDIBCO 2016 validation set for an ensemble of 5 FCNs in Table~\ref{tab:features} for the best performing parameterization of each feature type.

Relative Darkness (RD)~\cite{wu15} features performed best, though using the output of Howe's method~\cite{howe13} as an input feature performs almost as well.
RD is computed by counting the number of pixels in a local window that are darker, lighter, and similar to the central pixel.
We used RD features with a window size of 5x5 and a similarity threshold of $\pm10$ in all experiments in this paper.

\section{Conclusion}

In this work, we have proposed FCNs trained with a combined P-FM and FM loss for the task of document image binarization.
Our proposed method handles diverse domains of documents.
It out performs the competition winners for 4 of 7 DIBCO competitions and is competitive with the state-of-the-art on Palm Leaf Manuscripts.
We analyzed the architecture and found that performance is stable wrt changes in the architecture.
We found that the number of training images (i.e. diverse training data) is a more important factor than number of pixels used in training.
Finally, we analyzed using additional features as input to the FCN and found that Relative Darkness features~\cite{wu15} and the output of Howe binarization~\cite{howe13} perform best.

{\tiny
\newcommand{\BIBdecl}{\setlength{\itemsep}{0.25 em}}
\bibliographystyle{IEEEtran}
\bstctlcite{IEEEexample:BSTcontrol}
\bibliography{bib}

}
	
\end{document}